# Dilated Strip Attention Network for Image Restoration

Fangwei Hao[1], Jiesheng Wu, Ji Du, Yinjie Wang
Jing Xu, Member, IEEE

*Abstract*—Image restoration is a long-standing task that seeks to recover the latent sharp image from its deteriorated counterpart. Due to the robust capacity of self-attention to capture long-range dependencies, transformer-based methods or some attention-based convolutional neural networks have demonstrated promising results on many image restoration tasks in recent years. However, existing attention modules encounters limited receptive fields or abundant parameters. In order to integrate contextual information more effectively and efficiently, in this paper, we propose a dilated strip attention network (DSAN) for image restoration. Specifically, to gather more contextual information for each pixel from its neighboring pixels in the same row or column, a dilated strip attention (DSA) mechanism is elaborately proposed. By employing the DSA operation horizontally and vertically, each location can harvest the contextual information from a much wider region. In addition, we utilize multi-scale receptive fields across different feature groups in DSA to improve representation learning. Extensive experiments show that our DSAN outperforms state-of-the-art algorithms on several image restoration tasks.

## I. INTRODUCTION

With its goal of reconstructing a high-quality image from an observation that has suffered from various degradations such as blur, snowflake or haze, image restoration is crucial in numerous domains, including surveillance [1], [2], [3], medical imaging [4], [5], [6], and remote sensing [7], [8], [9]. It is an inverse issue with an ill-posed nature. To handle this difficult challenge, many conventional algorithms based on handcrafted features have been proposed, but they are inapplicable in more complex real-world scenarios [10]. Due to the strong learning ability, convolutional neural networks (CNNs) have significantly accelerated the progress of image restoration in recent years and demonstrated remarkable performance when compared to conventional methods [11], [12], [13]. By designing or proposing advanced units, numerous CNN-based techniques such as the U-shaped backbone [14], residual connection [15], dilated convolution [16], and attention modules [17], [18], have been proposed for a variety of image restoration tasks. However, convolutional neural network (CNN) with the static filters of the convolution operator cannot favorably handle the dynamic and non-uniform blur which is common in image restoration tasks. Besides, the convolution filter's small receptive field prevents it from simulating interactions between far-off pixels for large-scale blur. Despite numerous attempts to use dilated convolution [16] or stack deep layers to increase the receptive field, existing CNN modules encounters limited receptive field or abundant parameters.

Recently, state-of-the-art results on challenging high-level vision tasks [19], [20], [21] have been demonstrated by the transformer models which are derived from natural language processing. Through its core component, i.e., the self-attention mechanism, the transformer models can achieve effective modeling of long-range dependencies, which greatly improve the model performance. However, image restoration which inevitably involves high-resolution images, is relatively unfeasible due to its quadratic complexity with respect to spatial resolution of images. In order to address this problem, numerous methods have been proposed to increase image restoration's efficiency. Specifically, some networks [22], [23] limit the spatial zone of self-attention operation for simplification. In addition, instead of using self-attention among the spatial dimension, Restormer [24] utilizes it among the channel dimension so that it can capture long-range pixel interactions more efficiently. For the task of image deblurring, Stripformer [25] applies a novel strip-type self-attention operation which consists of intra- and inter-strip tokens to reweight image features in the horizontal and vertical directions. While these methods do provide a certain degree of complexity reduction, they do not change the intrinsic nature of self-attention; that is, network's complexity remains quadratic with respect to the size of windows, channels, or strips. More recently, to gather contextual information more efficiently for image restoration, Cui et al. [26] propose a strip attention module (SAM) to aggregate information from neighboring pixels in the same horizontal or vertical direction for each pixel. Although such a strip attention module can perceive information from an expanded region, it still encounters limited receptive field, resulting in capturing insufficient context information.

To address these issues, we propose a dilated strip attention network (DSAN) for image restoration in this paper. Specifically, we first elaborately design a dilated strip attention (DSA) mechanism to efficiently increase the strip receptive field, and it brings no additional parameter increase than the strip attention (SA) unit in [26]. By applying the DSA mechanism in the same horizontal and vertical direction for each pixel, we futhermore propose a dilated strip attention module (DSAM) to significantly enlarge square receptive field so that it can aggregate more contextual information from long-range neighboring pixels in the same horizontal and vertical direction for each pixel. The learnable weights, which are dynamically learned from the input feature by convolutional layers, are involved in this procedure. In order to improve the learning of feature representation, we also experimentally employ multi-scale receptive fields in feature groups to handle degrading blurs with different sizes.

[1]Fangwei Hao. is with the College of Artificial Intelligence, Nankai University, Tianjin, China (email: haofangwei@mail.nankai.edu.cn).
Jiesheng Wu. is with the College of Artificial Intelligence, Nankai University, Tianjin, China (email: jasonwu@mail.nankai.edu.cn).
Ji Du is with the College of Artificial Intelligence, Nankai University, Tianjin, China (email: 1120230244@mail.nankai.edu.cn).
Yinjie Wang is with the College of Artificial Intelligence, Nankai University, Tianjin, China (email: wangyinjie@mail.nankai.edu.cn).
Jing Xu is with the College of Artificial Intelligence, Nankai University, Tianjin, China (email: xujing@nankai.edu.cn).

Several significant benefits come with our DSAM. First of all, it achieves an extremely large receptive field with no parameter addition than the SAM. Furthermore, in contrast to the static convolutional filters, it is capable of processing different inputs and blurs via the learnable weights, which are dynamically learned from the input feature by convolutional layers. Thirdly, it can capture multi-scale contextual information.

When equipped with the proposed DSAM, our DSAN outperforms state-of-the-art algorithms on a number of image restoration tasks including image dehazing, motion deblurring and desnowing.

The main contributions of the paper are as follows:
- We first elaborately design a novel dilated strip attention (DSA) mechanism to efficiently enlarge the strip receptive field. By applying the DSA units in the same horizontal and vertical direction for each pixel, we futhermore propose a dilated strip attention module (DSAM) for image restoration to significantly enlarge square receptive field while multi-scale receptive fields are utilized across different feature groups to improve representation learning.
- By incorporating the proposed DSAM into a U-shaped backbone, we propose a dilated strip attention network (DSAN) for effective and efficient image restoration.
- Extensive experiments show that our DSAN outperforms previous state-of-the-art methods on several image restoration tasks.

## II. RELATED WORK

*A. Image Restoration*

Image restoration has drawn a lot of attention from the academic and industrial communities since it is crucial to photography, self-driving technologies, and medical imaging. In terms of its inverse issue with an ill-posed nature, numerous traditional methods have been established to limit the solution space, utilizing hand-crafted features, and a variety of assumptions [33]. Recently, the performance of image restoration has been greatly improved by deep CNN-based algorithms[34], [35], [36]. A general architecture for hierarchical feature representation learning in these networks is the U-shaped network [37]. In addition, a great deal of sophisticated modules such as various attention mechanisms [17], universal skip connection [38], and effective dilated convolution [16], have been proposed or introduced from high-level tasks. Besides, in order to better capture long-range dependencies, transformer-based models [22] have been introduced into low-level tasks. More recently, a strip attention module (SAM) [26] is proposed to learn contextual information from an expanded region, and it boosts the model performance on several image restoration tasks.

*B. Attention Mechanism*

In the field of computer vision, attention mechanism has been used extensively. Massive attention modules have been proposed for image restoration with the goal of capturing interdependencies along channels [39], [24], spatial coordinates [40], or both [41]. For example, FFA-Net [17] uses pixel and channel attention to handle various kinds of information in a flexible manner. By using channel-wise attention, Liu et al. [39] propose a GridDehazeNet to adjust the contributions of different feature maps to effectively fuse feature. Besides, the supervised attention module in MPRNet [42] is utilized for feature filtering. The performance of image restoration tasks has been promoted thanks to these attention modules.

The development of effective self-attention for image restoration is a successive research field. In particular, similar to Swin Transformer [43], Uformer [23] and SwinIR [22] both leverage the effective self-attention mechanism within local regions. Besides, Restormer [24] pays attention to channel self-attention rather than spatial dimensions. To remove motion blur, interlaced intra-strip and inter-strip attention layers are developed by Stripformer [25] to enhance model performance. Nevertheless, as for the size of the region, channel or strip, the complexity of these remedies is still quadratic. In addition, in order to learn contextual information from an expanded region, a strip attention module (SAM) [26] is proposed, and it improves the model's performance on several image restoration tasks. Yet the SAM still encounters limited receptive field, resulting in capturing insufficient context information.

In this paper, we first elaborately design a novel dilated strip attention (DSA) mechanism to efficiently enlarge the strip receptive field. We futher propose a novel dilated strip attention module (DSAM) which sequentially conducts effective horizontal and vertical information integration. In addition to maintaining high efficiency, it yields dynamic aggregation weights and a much larger receptive field as compared to the standard convolution operator.

## III. METHODOLOGY

This section first presents the architecture of the proposed DSAN for image restoration tasks. We then show the proposed DSAM in detail after describing the novel dilated strip attention (DSA) operation. At last, we detail the loss functions for training.

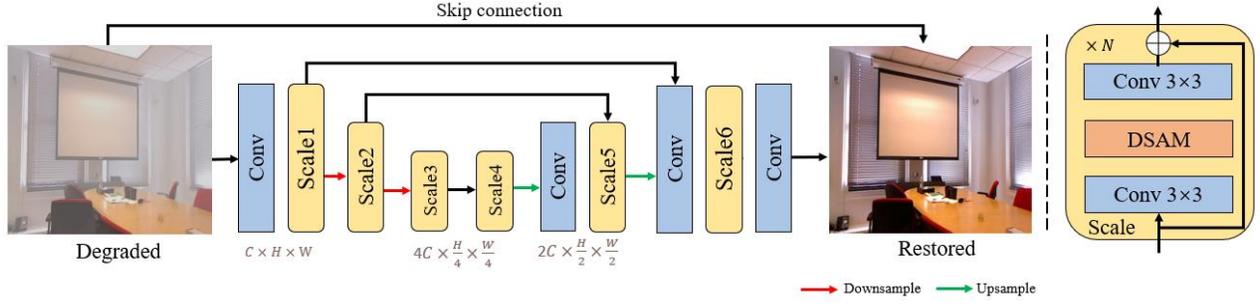

Figure 1. Network architecture of our DSAN.

*A. Network Architecture*

In order to learn hierarchical representations effectively, our DSAN applies the widely used encoder-decoder architecture and consists of six scales totally. Figure 1 shows the overall pipeline of the proposed DSAN. Specifically, a single convolution layer is used to extract shallow feature maps with size of $R^{H \times W \times C}$ from an input degraded image with the size of $R^{3 \times H \times W}$. The obtained shallow feature is then fed into the encoder layers (Scale 1-3) for feature refining. During this process, the spatial resolution is gradually decreased from $R^{H \times W \times C}$ to $R^{4C \times \frac{H}{4} \times \frac{W}{4}}$, while the number of channels is increased stage by stage. At each stage, several residual blocks are stacked for effective feature learning, followed by the last one residual block which involves the proposed DSAM, and the strided convolution then performs the downsampling operation. Afterwards, the lowest-resolution feature maps are fed into the decoder layers (Scale 4-6) in order to gradually recover the high-resolution representations. As for the feature upsampling, we use the transposed convolution to generate high-resolution feature maps. Similar to previous works [23], [24], the feature-level skip connections are applied to mitigate the problem of information loss caused by downsampling operation. After the encoder features and corresponding decoder features are concatenated, a convolution layer is utilized to adjust the channel dimension. Finally, The original input image is added to create the final sharp image, forcing the network to concentrate only on the residual information learning. Additionally, following previous methods [15], [44], we also adopt multi-input and multi-output strategies to lessen the difficulty of training.

*B. Dilated Strip Attention Module (DSAM)*

Our primary aim is to design an effective and efficient attention module for integrate information. We first provide self-attention complexity analyses and strip attention module (SAM) [26] for revisiting. Then, we go on to describe the formulation of the proposed dilated strip attention.

**Self-Attention**

High-level vision tasks have been demonstrated great success with self-attention. However, image restoration tasks which always deal with high-resolution images are unfeasible due to the quadratic complexity of self-attention module. As for self-attention, suppose that $\mathbf{X} \in R^{H \times W \times C}$ denotes the input tensor, where $H \times W$ and C are the spatial resolution and the number of channels, respectively. The self-attention can be formulated as

$$\text{Attention}(\mathbf{Q}, \mathbf{K}, \mathbf{V}) = \text{Softmax}(\mathbf{QK}^\top)\mathbf{V}, \text{ where } \mathbf{Q} = \mathbf{XW}^Q, \mathbf{K} = \mathbf{XW}^K, \mathbf{V} = \mathbf{XW}^V$$

where $\mathbf{Q}, \mathbf{K},$ and $\mathbf{V} \in R^{HW \times C}$, and they are generated by utilizing corresponding projection matrices ($\mathbf{W}^Q, \mathbf{W}^K,$ and $\mathbf{W}^V$) and reshaping, respectively. For simplicity, we leave out the normalization term. In terms of Attention($\mathbf{Q}, \mathbf{K}, \mathbf{V}$), we can see that the complexity of the calculation of $\mathbf{Q}, \mathbf{K}, \mathbf{V}$ is $3HWC^2$. Based on $\mathbf{Q}$ and $\mathbf{K}$, the dot-product to generate the attention map has the complexity of $(HW)^2 C$. Besides, the complexity of the weighted summation process is $(HW)^2 C$. It is obvious that the last two complexities are quadratic with respect to the size of spatial resolution.

**Dilated Strip Attention**

Before introducing dilated strip attentionin (DSA) in detail, let us first review the details of strip attention (SA) in [26]. To reduce the calculation complexity of self-attention, Cui et al. [26] devise an effective operator to aggregate information through a strip attention unit. For the horizontal strip attention operator, after the input feature $\mathbf{X} \in R^{H \times W \times C}$ is given, the global average pooling (GAP) is used to calculate the spatial information followed by a $1 \times 1$ standard convolution layer and sigmoid function to yield the attention weights. This process can be formulated as

$$\mathbf{A} = \sigma(W_{1 \times 1}(\text{GAP}(\mathbf{X})))$$

where $W_{1\times1}(\cdot)$ and $\sigma(\cdot)$, $\mathbf{A}$ are the $1 \times 1$ standard convolution layer, the Sigmoid function, and the obtained attention weights, respectively. Besides, $\mathbf{A} \in R^K$ and $K$ is the integration strip's length. Then, the weighted sum is operated within the strip of size $1 \times K$, and a convolution-type integration is adopted based on the attention weights $\mathbf{A}$. The process can be formally expressed as

$$\widehat{\mathbf{X}}_{h,w,c} = \sum_{k=0}^{K-1} \mathbf{A}_k \mathbf{X}_{h, w-\lfloor\frac{K}{2}\rfloor+k, c}$$

As a result, such strip attention operation greatly simplifies the calculation of self-attention and achieves excellent performance at the same time. In order to simplify the expression, the process can be formally expressed as

$$\widehat{\mathbf{X}} = F_{SA}^K(\mathbf{X})$$

Where $F_{SA}^K(\cdot)$ denotes the strip attention (SA) operation, and $K$ is the size of the strip.

However, using the SA operator, the receptive field of each new pixel is $K$, which is the fixed strip size. On the other hand, the dilation operation with no parameter increase in [45] is an effective strategy to increase the receptive field, and their proposed dilation convolutions, whose receptive fields are a square of exponentially increasing size, can capture more contextual information. Motivated by this fact, we further develop a novel dilated strip attention (DSA) by integrating SA and the dilation operation to enlarge the direct receptive field for capturing more contextual information. The process of DSA can be formally expressed as

$$\widehat{X}'_{h,w,c} = \sum_{k=0}^{K-1} A_k X_{h,w+\left(k-\left\lfloor\frac{K}{2}\right\rfloor\right)*d,c}, d = 1, \dots, D,$$

Where $d$ is the the dilation operation rate, $K$ is the size of the strip. It can be seen that the DSA degenerates to SA when the dilation rate $d$ is set to 1. Moreover, it is worth noting that our DSA with learned attention weights $A_k$ does not increase additional module parameters compared to SA. Likewise, in order to simplify the expression, the process of DSA can be formally expressed as

$$\widehat{X}' = F_{DSA}^K(X)$$

Where $F_{DSA}^K(\cdot)$ denotes the strip attention (DSA) operation, and $K$ is the size of the strip.

## C. Dilated Strip Attention Module (DSAM)

Based on strip attention (SA) [26], Cui et al. proposed the strip attention module (SAM) by ultilizing SA operation horizontally and vertically, as shown in Figure 2. Although SAM can perceive information from an expanded region, it still encounters limited square receptive field, resulting in capturing insufficient context information.

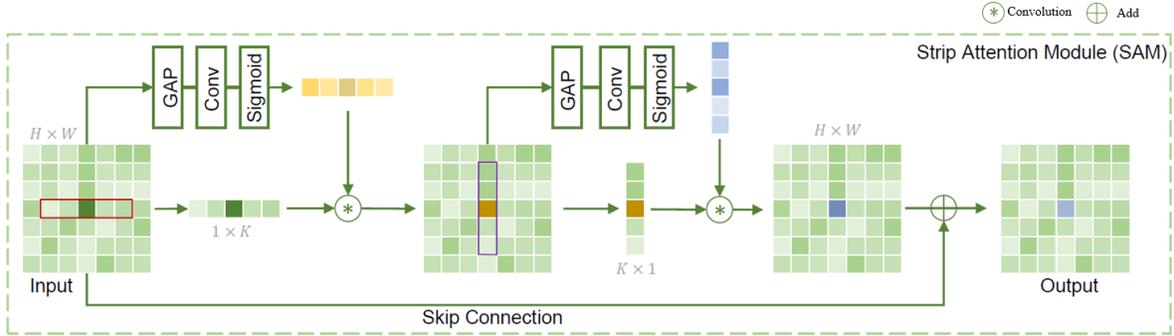

Figure 2. The details of the SAM for revisiting.

To further enlarge the receptive field and harvest more contextual information, we propose a dilated strip attention module (DSAM) by carrying out DSA operations in both vertical and horizontal directions. Compared to SAM, the proposed DSAM with no parameters increase can capture more context information, leading to boosting the performance of image restoration tasks. The details of the DSAM are shown in Figure 3 when the dilation rate $d$ is set to 2, and we can adjust the size of the receptive field by changing the value of dilation rate. When the value of dilation is set to 1, DSAM degenerates into the regular SAM. Therefore, DSAM is a general form of SAM while SAM is a special form of DSAM. When equipped with the proposed DSAM, thanks to the larger receptive field obtained, the DSAN can learn more contextual information which helps to improve the model performance.

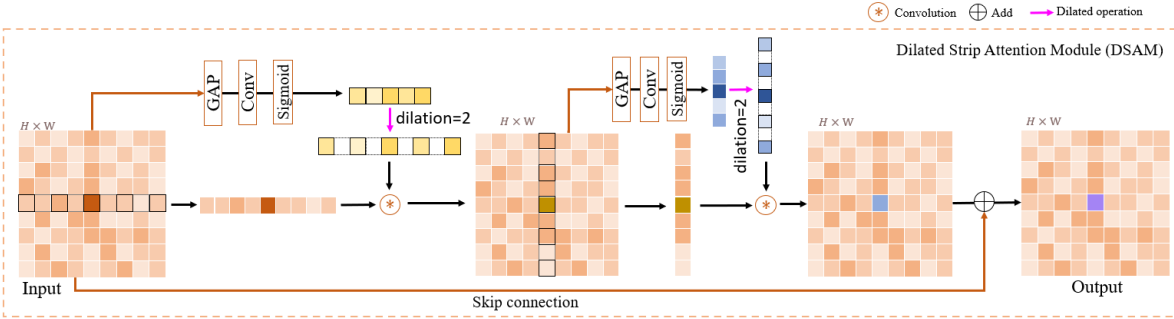

Figure 3. The details of the proposed DSAM.

## D. Loss Functions

Following [15], We employ the dual-domain $L_1$ loss to enable feature refinement in both spatial and frequency domains for our network training. The loss function for each output can be expressed as

$$L_s = \frac{1}{S} \| \hat{\mathbf{I}} - \mathbf{I} \|,$$

$$L_f = \frac{1}{S} \| \mathcal{F}(\hat{\mathbf{I}}) - \mathcal{F}(\mathbf{I}) \|,$$

$$L = L_s + \lambda L_f$$

where $\mathbf{I}$ and $\hat{\mathbf{I}}$ denote the ground truth and the predicted image, respectively; $\mathcal{F}(\cdot)$ is the fast Fourier transform (FFT), and $S$ represents the total elements for normalization. During training process, $\lambda$ is set to 0.1, and we adopt the Adam optimization algorithm [46] to optimize the entire network.

## IV. EXPERIMENTAL RESULTS

In this section, to verify our DSAN's effectiveness, extensive experiments are conducted on several image restoration tasks, including image motion deblurring (GoPro [29] and HIDE [30]), image dehazing (RESIDE [28]), and image desnowing (CSD [32]). An explanation of our results based on these datasets comes after the presentation of the training settings. Then, a series of ablation experiments are carried out. Lastly, the model complexity analysis is carried out.

### A. Settings

The Adam [46] optimizer with $\beta_1 = 0.9$ and $\beta_2 = 0.999$ is adopted to optimize the proposed network. We set the initial learning rate at $1e^{-4}$, and it decays to $1e^{-6}$ with the cosine annealing strategy. For the RESIDE-Outdoor [28] dataset, we set the batch size to 8, and for other datasets, the batch size is set to 4. During training, models are trained using a 256×256 patch size. For data augmentation, we only use horizontal flips. In addition, we use different numbers of residual blocks $N$ in each scale for different tasks, e.g., $N = 4$ for image dehazing and desnowing, and $N = 16$ for image defocus deblurring, depending on the complexity of the task.

### B. Main Results

**Image dehazing.** The RESIDE [28] dataset is utilized to train our DSAN, and the SOTS [28] dataset is used for testing. Table 1 presents the testing results. Compared to state-of-the-art methods including DCP [47], GCANet [48], GridDehazeNet [39], MSBDN [49], PFDN [50], FFA-Net [17], AECR-Net [51], MAXIM [52], DeHamer [53], PMNet [54], DehazeFormer-L [55] and SANet [26], our DSAN yields the highest quantitative results and outperforms the recent algorithm DehazeFormer-L by 0.55 dB PSNR, reaching 40.60 dB on the SOTS-Indoor dataset. It is worth noting that after adopting the proposed DSAM module, the PSNR performance of our DSAN increases by 0.2 dB and no additional parameters are increased when compared with SANet. In terms of SOTS-Outdoor dataset, our DSAN with less complexity achieves the highest performance of 38.41 dB PSNR, which demonstrates the superiority of our method. Specifically, with just 76% MACs and 3% parameters of the expensive Transformer model DeHamer [53], our DSAN outperforms it by 3.23 dB PSNR. Besides, Our DSAN without any parameter increase achieves a 0.40 dB PSNR gain and 0.001 SSIM gain compared to SANet. As for the visual comparison, Figure 4 shows the visual and corresponding quantitative results of different methods on "1412_10.png" of SOTS-Indoor dataset. Visual results show that our DSAN is better at dehazing and producing sharper details, and the corresponding quantitative results are consistent with the same conclusion. In addition, figure 5 presents the comparisons of different methods on the "1869_1_0.2.png" of SOTS-Outdoor dataset. Its visual and quantitative results reveal the similar findings to the conclusion on the SOTS-Indoor dataset. Experimental results show that our DSAN can yield clearer images which are visually closer to the reference images than other algorithms, and we can see that our DSAN is more effective and efficient when applied to image dehazing task.

Table 1: Image dehazing results on SOTS [Li *et al.*, 2018]. DSAN receives higher scores with fewer Params than most competitors.

| Method | SOTS-Indoor | | SOTS-Outdoor | | Overhead | |
|---|---|---|---|---|---|---|
| | PSNR↑ | SSIM↑ | PSNR↑ | SSIM↑ | Params (M) | MACs (G) |
| DCP | 16.62 | 0.818 | 19.13 | 0.815 | - | - |
| GCANet | 30.23 | 0.980 | - | | 0.702 | 18.41 |
| GridDehazeNet | 32.16 | 0.984 | 30.86 | 0.982 | 0.956 | 21.49 |
| MSBDN | 33.67 | 0.985 | 33.48 | 0.982 | 31.35 | 41.54 |
| PFDN | 32.68 | 0.976 | - | | 11.27 | 50.46 |
| FFA-Net | 36.39 | 0.989 | 33.57 | 0.984 | 4.456 | 287.8 |
| AECR-Net | 37.17 | 0.990 | - | | 2.611 | 52.20 |
| MAXIM | 38.11 | 0.991 | 34.19 | 0.985 | 14.1 | 108 |
| DeHamer | 36.63 | 0.988 | 35.18 | 0.986 | 132.45 | 48.93 |
| PMNet | 38.41 | 0.990 | 34.74 | 0.985 | 18.90 | 81.13 |
| DehazeFormer-L | 40.05 | **0.996** | - | | 25.44 | 279.7 |
| SANet | 40.40 | **0.996** | 38.01 | 0.995 | 3.81 | 37.26 |
| DSAN (Ours) | **40.60** | 0.996 | **38.41** | 0.996 | 3.81 | 37.26 |

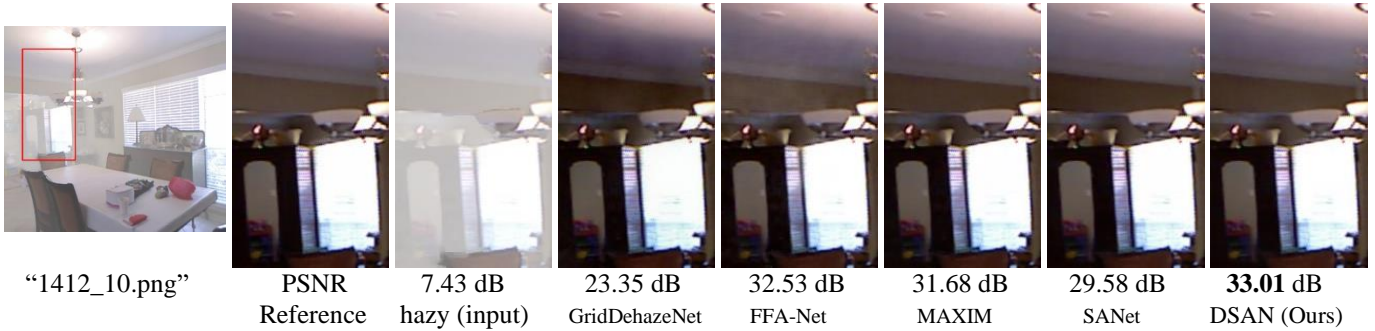

Figure 4. Comparisons of Image dehazing on the SOTS-Indoor. The best result is **highlighted**.

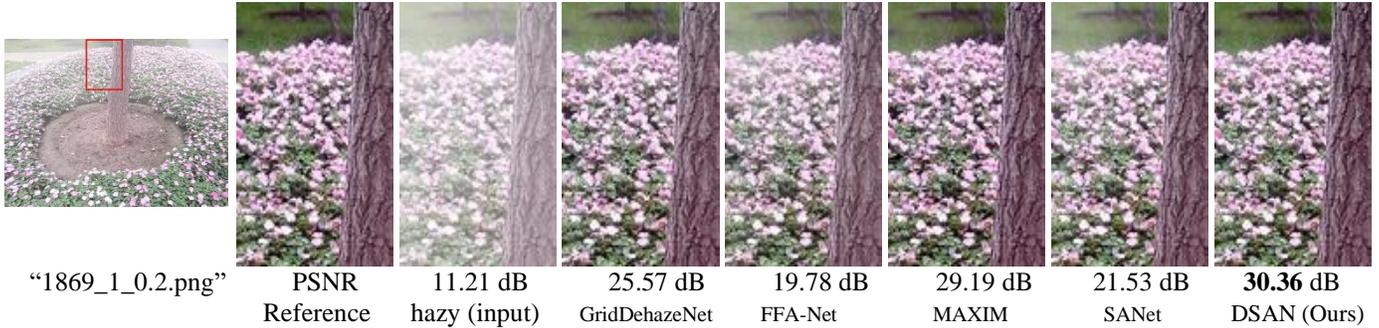

Figure 5. Comparisons of Image dehazing on the SOTS-Outdoor. The best result is **highlighted**.

**Image motion deblurring.** In order to have a more comprehensive validation of the proposed algorithm, the synthetic datasets are used to evaluate our DSAN. Specifically, we train the model on the GoPro training set and validate the model on the testing dataset of GoPro, and we also test the performance of the obtained model on the HIDE dataset. We compare the PSNR values of our method with the results of previous state-of-the-art methods for single-image motion deblurring task, including DeblurGAN-v2 [56], DBGAN [57], DMPHN [58], SPAIR [59], MIMO-UNet+ [60], IPT [61], MPRNet [42], HINet [62], and Restormer[24]. Table 2 presents detailed comparison results, which indicate that our model performs better than the powerful transformer model Restormer [24]. Specifically, our DSAN outperforms Restormer by 0.01 dB PSNR on the GoPro testing set, and the obtained 0.966 SSIM of our DSAN are higher than the 0.961 SSIM of Restormer. In addition, in terms of HIDE dataset, although the 30.85 dB PSNR obtained by our method is lower than the 31.22 dB PSNR of the Restormer, our obtained 0.953 SSIM is 0.011 higher than the 0.942 SSIM of the Restormer, indicating that our DSAN is able to better recover the structure of the objects in the image. Visual results are further illustrated in Figure 6. In comparison to other competing frameworks, the visual results indicate that the proposed DSAN can recover more faithful output, generating sharper contours and more detailed information. Such results are due to more contextual information extracted from the larger receptive field.

Table 2: Image motion deblurring results on the GoPro and HIDE datasets, the best and the second best results are **hightlighted** and <u>underlined</u>, respectively.

| Method | GoPro | | HIDE | |
|---|---|---|---|---|
| | PSNR↑ | SSIM↑ | PSNR↓ | SSIM↓ |
| DeblurGAN-v2 | 29.55 | 0.934 | 26.61 | 0.875 |
| DBGAN | 31.10 | 0.942 | 28.94 | 0.915 |
| DMPHN | 31.20 | 0.940 | 29.09 | 0.924 |
| SPAIR | 32.06 | 0.953 | 30.29 | 0.931 |
| MIMO-UNet+ | 32.45 | 0.957 | 29.99 | 0.930 |
| IPT | 32.52 | - | - | - |
| MPRNet | 32.66 | 0.959 | 30.96 | 0.939 |
| HINet | 32.71 | 0.959 | 30.32 | 0.932 |
| Restormer | <u>32.92</u> | <u>0.961</u> | **31.22** | <u>0.942</u> |
| DSAN (Ours) | **32.93** | **0.966** | <u>30.85</u> | **0.953** |

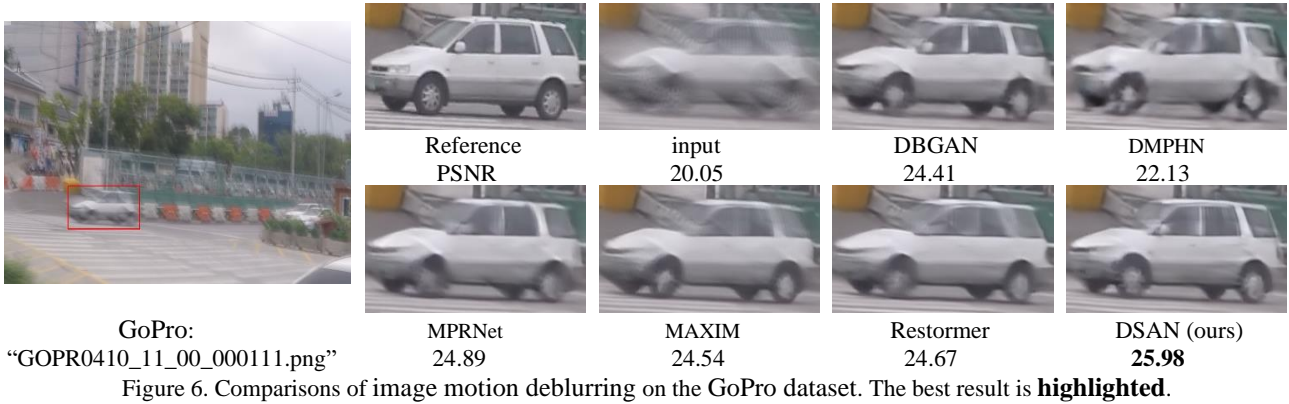

Figure 6. Comparisons of image motion deblurring on the GoPro dataset. The best result is **highlighted**.

**Image desnowing.** Table 3 presents the desnowing comparisons on the CSD [32] dataset. It can be seen that our method outperforms other state-of-the-art methods, including DesnowNet [63], CycleGAN [64], All in One [65], JSTASR [66], HDCW-Net [32], TransWeather [67], MSP-Former [41], NAFNet [31] and SANet [26], in terms of PSNR and SSIM. In contrast to the latest algorithms NAFNet and SANet, our DSAN achieves 1.43 dB and 0.17 dB PSNR gain, respectively. In addition, our model outperforms the Transformer model MSP-Former by 2.81 dB PSNR and 0.025 SSIM. Visual results in Figure 7 demonstrate that our DSAN can remove smaller snowflakes, resulting in clearer and more natural visual results than competing algorithms.

Table 3: Image desnowing results on CSD [Chen et al., 2021]. DSAN outperforms other methods significantly.

| Method | PSNR | SSIM |
|---|---|---|
| DesnowNet | 20.13 | 0.81 |
| CycleGAN | 20.98 | 0.80 |
| All in One | 26.31 | 0.87 |
| JSTASR | 27.96 | 0.88 |
| HDCW-Net | 29.06 | 0.91 |
| TransWeather | 31.76 | 0.93 |
| MSP-Former | 33.75 | 0.96 |
| NAFNet | 35.13 | 0.97 |
| SANet | 36.39 | 0.98 |
| DSAN (Ours) | **36.56** | **0.985** |

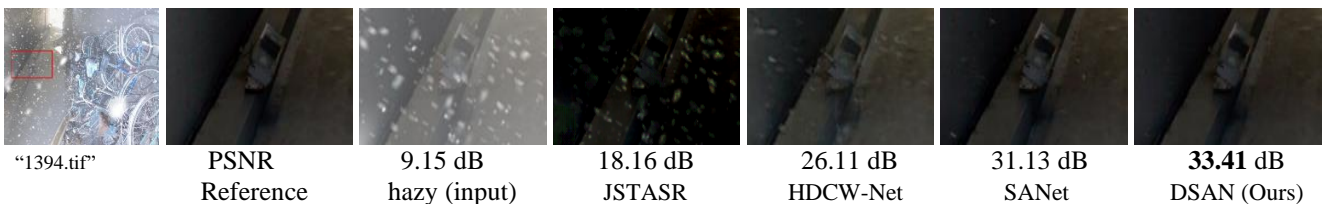

Figure 7. Comparisons of Image desnowing on the CSD dataset. The best result is **highlighted**.

*C. Ablation studies*

In the ablation experiment, we record and analyze the performance of the model with different modules. First, we use the model without DSAM as the basic model to train on the RESIDE [28] dataset and test on the SOTS-Indoor [28], and we record the tested performance values of PSNR and SSIM as the performance benchmark.

Then, DSAM modules with different dilation rate, i.e., 1, 2, 3, 4, are respectively added to the basic model. Under the same training and testing conditions, they are trained and tested respectively, and the obtained model performance values of PSNR and SSIM are recorded, as shown in Table 4.

It can be seen that our DSAN with DSAM achieves better performance values of PSNR and SSIM than the basic model without DSAM. For DSAM modules with different dilation rates, our DSAN achieves the best performance when the dilation rate is set to 3, exceeding the model when the dilation rate is set to 2 by 0.13 dB PSNR, and outperforms the model with the dilation rate 1 by 0.20 dB PSNR. It is worth noting that the number of parameters does not vary when DSAM modules are set to different dilation rates. The experimental results show that the dilated strip attention mechanism is effective in boosting performance.

In addition, the obtained model performance when the dilation rate is 4 does not increase further than the performance when the dilation rate is set to 3, indicating that the performance of the DSAM module is not only affected by the size of the receptive field, but also by the amount of learnable parameters.

Table 4: Ablation studies on SOTS-Indoor dataset. The best result is **highlighted**.

| Method | PSNR | SSIM |
|---|---|---|
| Without DSAM | 38.19 | 0.995 |
| DSAM (dilation=1) | 40.40 | 0.996 |
| DSAM (dilation=2) | 40.47 | 0.996 |
| DSAM (dilation=3) | **40.60** | 0.996 |
| DSAM (dilation=4) | 40.40 | 0.996 |

*D. Model Complexity Analysis*

To further validate the efficiency of our DSAN, we analyze the complexity of the model and compare it with other previous state-of-the-art methods including DeHamer [53], MAXIM [52], PMNet [54], DehazeFormer-L [55] and SANet [26]. Specifically, we analyze and compare the number of model parameters, model's multi-adds and model performance on the SOTS-Indoor dataset, and we show the detailed comparison results in Table 5. It can be seen that our model achieves the best performance with the least number of parameters and multi-adds compared to recent methods. Notably, our DSAN has the same number of parameters and multi-adds as the SANet but achieves a performance gain of 0.2dB. Figure 8 further visually illustrates the difference between the different methods in terms of parameters and PSNR, and Figure 9 visually illustrates their difference about multi-adds and PSNR.

Table 5. Computation and parameters comparison for image dehazing (SOTS-Indoor dataset). The best results are **hightlighted.**

| Metric | DeHamer | MAXIM | PMNet | DehazeFormer-L | SANet | DSAN |
|---|---|---|---|---|---|---|
| Paras (M) | 132.45 | 14.1 | 18.90 | 25.44 | **3.81** | **3.81** |
| Mult-Adds (G) | 48.93 | 108 | 81.13 | 279.7 | **37.26** | **37.26** |
| PSNR (dB) | 36.63 | 38.11 | 38.41 | 40.05 | 40.40 | **40.60** |

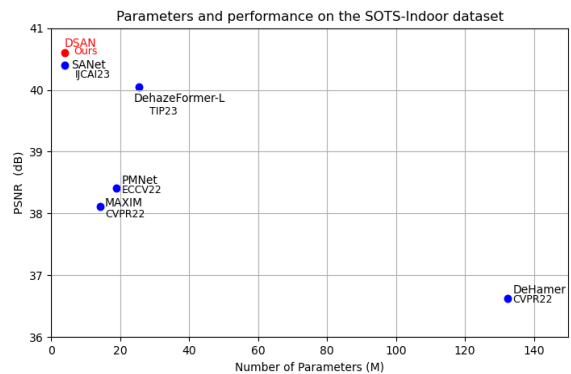

Figure 8. Performance and the parameters of different methods on SOTS-Indoor dataset.

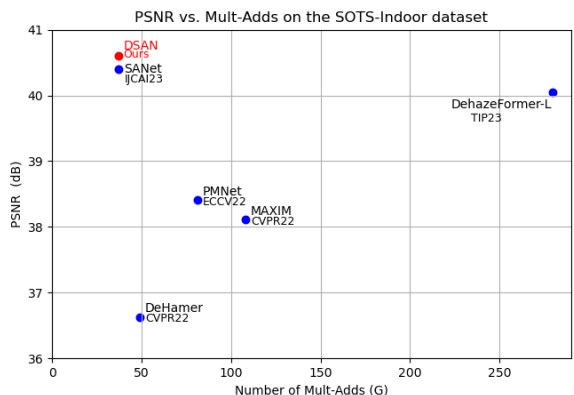

Figure 9. Performance and the Multi-Adds of different methods on SOTS-Indoor dataset.

V. CONCLUSION

In this paper, we propose a dilated strip attention network (DSAN) for effective and efficient image restoration by integrating more contextual information to improve feature representation. Specifically, we exploit a novel dilated strip attention (DSA) mechanism based on the learned attention weights to simulate interactions between far-off pixels for large-scale image degradation. Furthermore, we propose a dilated strip attention module (DSAM) by horizontally and vertically ultilizing the DSA mechanism. The DSAM with the content-aware property can maintain the advantages of self-attention in capturing more long-range dependencies due to the dilated receptive field. Moreover, the DSAM uses multi-scale kernels to learn vital multi-scale information to effectively deal with image degradation of different sizes. Extensive experiments show that our DSAN outperforms state-of-the-art algorithms on several image restoration tasks.

ACKNOWLEDGMENT

The authors acknowledge the anonymous reviewers for their helpful comments.